\documentclass[11pt]{article}

\usepackage{microtype} 
\usepackage{booktabs}  
\usepackage{multirow}
\usepackage{chngpage}
\usepackage{url}  

\usepackage{graphicx}

\usepackage{amsmath}
\usepackage{amsthm}
\DeclareMathOperator*{\argmin}{arg\,min}

\usepackage[final]{automl}

\usepackage{natbib}
\title{EmProx: Neural Network Performance Estimation\\For Neural Architecture Search}

\author[1]{\nameemail{G.G.H. Franken}{g.g.h.franken@student.tue.nl}}
\author[1]{\nameemail{P. Singh}{p.singh@tue.nl}}
\author[1]{\nameemail{J. Vanschoren}{j.vanschoren@tue.nl}}

\affil[1]{Eindhoven University of Technology}

\hypersetup{%
  pdfauthor={G.G.H. Franken, P. Singh, J. Vanschoren}, 
  pdftitle={EmProx: Neural Network Performance Estimation For Neural Architecture Search},
  pdfsubject={Neural Architecture Search},
  pdfkeywords={Neural Architecture Search, Performance Estimation, NAO}
}

\begin{document}

\maketitle

\begin{abstract}
Common Neural Architecture Search methods generate large amounts of candidate architectures that need training in order to assess their performance and find an optimal architecture. To minimize the search time we use different performance estimation strategies. The effectiveness of such strategies varies in terms of accuracy and fit and query time. This study proposes a new method, EmProx Score (Embedding Proximity Score). Similar to  Neural Architecture Optimization (NAO), this method maps candidate architectures to a continuous embedding space using an encoder-decoder framework. The performance of candidates is then estimated using \emph{weighted kNN} based on the embedding vectors of architectures of which the performance is known. Performance estimations of this method are comparable to the MLP performance predictor used in NAO in terms of accuracy, while being nearly nine times faster to train compared to NAO. Benchmarking against other performance estimation strategies currently used shows similar to better accuracy, while being five up to eighty times faster. 

Code is made publicly available on GitHub\footnote{https://github.com/GideonFr/EmProx}.

\end{abstract}

\section{Introduction}
Neural architecture search (NAS) is the science of automatically designing novel neural network architectures for a specific task. The NAS problem is described as follows in \cite{White2021}: given architectures $a$ from search space $\mathcal{A}$, find the optimal architecture $a^*$ as $a^* = \argmin_{a\in\mathcal{A}}f(a)$, where $f$ denotes the validation error of architecture $a$ on a fixed dataset for a fixed number of epochs $\mathit{E}$. However, evaluating $f(a)$ requires significant computational resources. Hence, performance predictor $f'$ is leveraged, defined as any function that predicts $f(a)$ without fully training $a$. Evaluating $f'$ should take significantly less time than evaluating $f$ and desirably $\{f'(a) \mid a\in\mathcal{A}\}$ has a high correlation with $\{f(a) \mid a\in\mathcal{A}\}$ 

We propose a new performance predictor $f'$ called EmProx Score (Embedding Proximity Score) as a faster alternative for Neural Architecture Optimization (NAO) \citep{Luo2018}. Similar to NAO, EmProx uses an encoder-decoder framework to map candidate architectures to a continuous embedding space. However, instead of training an multi-layer perceptron (MLP) on the embedding vectors to predict the accuracy of a candidate, EmProx uses \emph{weighted k-nearest neighbors} on the embedding vectors. As such, EmProx estimates the performance of the query architecture based on the distance and accuracy of its nearest neighbors in the embedding space, which can be computed efficiently.

In the remainder of this paper, we first describe background and related work in Section \ref{sec:relwork}. Section \ref{sec:methods} describes the EmProx procedure in more detail, and Section \ref{sec:design} discusses implementation and evaluation details. Section \ref{sec:experiments} evaluates this method against a wide range of alternative approaches, and Section \ref{sec:conclusions} concludes.

\section{Related work}
\label{sec:relwork}
\begin{figure}[tp]
\centering
\begin{tabular}{r}
\includegraphics[width=1\textwidth]{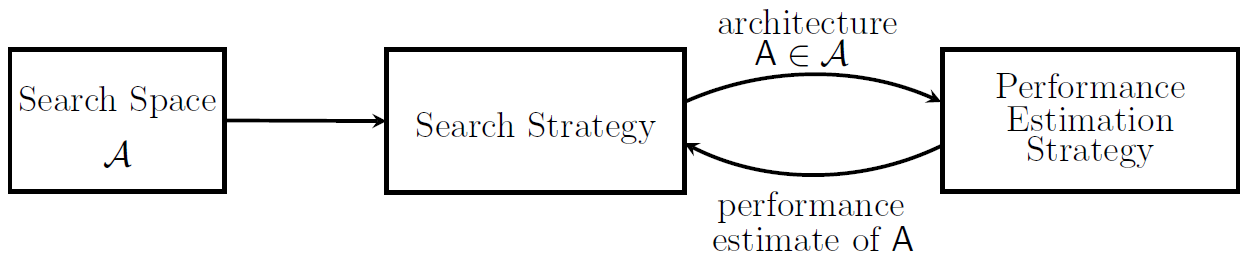}
\end{tabular}
\caption{The three components of a Neural Architecture Search algorithm. The search space defines the model architectures that can be discovered. The search strategy selects architectures from the search space to evaluate. To evaluate an architecture, a performance estimation strategy is used. Usually, this strategy returns the expected architecture's accuracy to the search strategy to further guide the architecture search.}
\label{fig:components}
\end{figure}

NAS methods consist of three components: the search space, the search strategy/(optimizer) and the performance estimation strategy \citep{Elsken2018}. Figure \ref{fig:components} shows how these components relate to each other. 

\subsection{Differentiable architecture search}
The search space describes the way architectures are defined and what architectures can be discovered. A commonly used approach is a cell search space, in which small networks (cells) are learned which are then repeated in a macro-architecture. This discrete search space can also be made continuous by defining a zero-shot architecture, a super-network consisting of all possible operations, and then applying continuous relaxation, adding a continuous meta-parameter for every possible option, as is done in DARTS \citep{Liu2018}.

The search strategy defines how the search space is explored to select candidate architectures, usually leveraging performance estimations of previous candidates to select more promising candidates. Common strategies include evolutionary algorithms \citep{DBLP:journals/corr/RealMSSSLK17} and reinforcement learning methods \citep{DBLP:journals/corr/BakerGNR16}. Gradient-based techniques can also be used: DARTS leverages weight sharing, optimizing both the meta-parameters and the model weights at the same time using bilevel optimization, thus evaluating multiple architectures simultaneously. While this is very efficient, the downside is that many performance estimation strategies are not applicable to DARTS.


\subsection{Neural Architecture Optimization}
\begin{figure}[htp]
\centering
\begin{tabular}{r}
\includegraphics[height=2.4in]{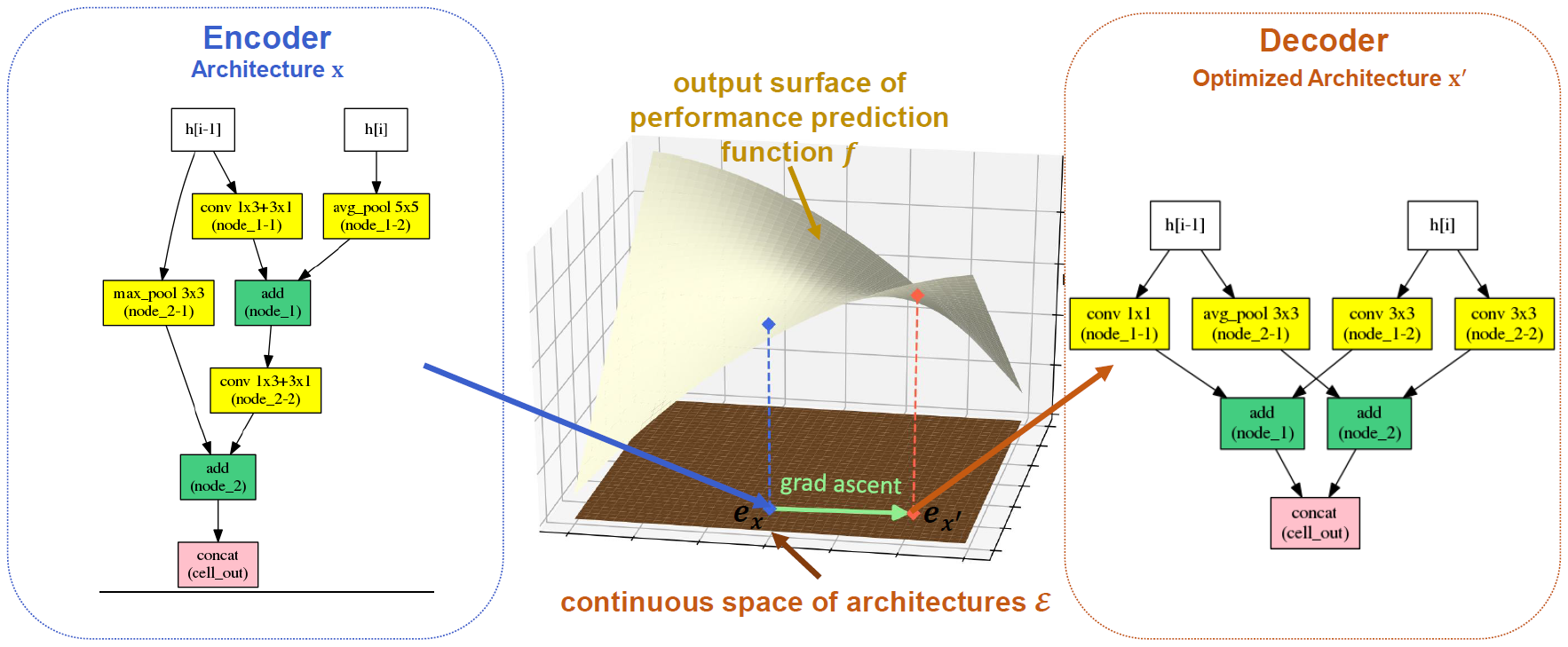}
\end{tabular}
\caption{The NAO framework. Architecture $x$ is mapped to a continuous embedding space by the encoder. Using its embedding $e_x$ and predictor $f$, its performance is predicted. Then the performance of $e_x$ is optimized into $e_x'$ by maximizing the output of predictor $f$ using gradient ascent. To get the improved architecture $x'$, the decoder maps $e_x'$ to the discrete space. From \cite{Luo2018}.}
\label{fig:NAO}
\end{figure}

A solution to this problem is offered by Neural Architecture Optimization (NAO), by \cite{Luo2018}. It maps a neural network architecture $x$ to an embedding space using an encoder-decoder model, as shown in Figure \ref{fig:NAO}. The output of the encoder is a continuous feature vector $e_x$ (i.e., an embedding), that captures the semantics of the architecture that served as input for the encoder. A performance predictor (NAO uses an MLP) then uses the embedding $e_x$ to predict the architecture's accuracy. The output of the performance predictor is then maximized using gradient ascent to find $e_x'$, the embedding of an improved architecture $x'$. To get architecture $x'$, the decoder maps $e_x'$ back to the discrete space. This framework, using a performance predictor in a continuous search space, allows NAO to perform gradient-based search in the embedding space quickly. An important limitation of NAO is that the performance predictor (MLP) first needs to be trained, which is a practical bottleneck. This study, as described in detail in section \ref{sec:methods}, proposes a modification of NAO that uses a performance estimation strategy that makes the MLP redundant. \\

\subsection{Performance estimation strategies}
Performance estimation strategies, or performance predictors, were developed since it is often infeasible in NAS to train every candidate architecture from scratch. NAO uses an MLP, but many alternatives exist. These are often divided into the following categories: 1) lower fidelity estimates, 2) learning curve extrapolation, 3) weight inheritance and 4) one-shot models/weight sharing. The first method reduces training times by training for fewer epochs, a subset of the data, downscaled models etc. The second method aims to extrapolate performance from a few epochs of training. In the third method, models are not trained from scratch, but initialized by inheriting weights from a parent model. The fourth category, used in DARTS, trains only a single large one-shot model, using the loss of the current one-shot model itself to optimize both the meta-parameters and model weights. Next to these methods, surrogate models can also be used to predict the performance of candidate pipelines. While NAO uses an MLP, we will leverage \emph{weighted kNN} as a surrogate model.

\section{Methods}
\label{sec:methods}
\begin{figure}[htp]
\centering
\begin{tabular}{r}
\includegraphics[height=1.9in]{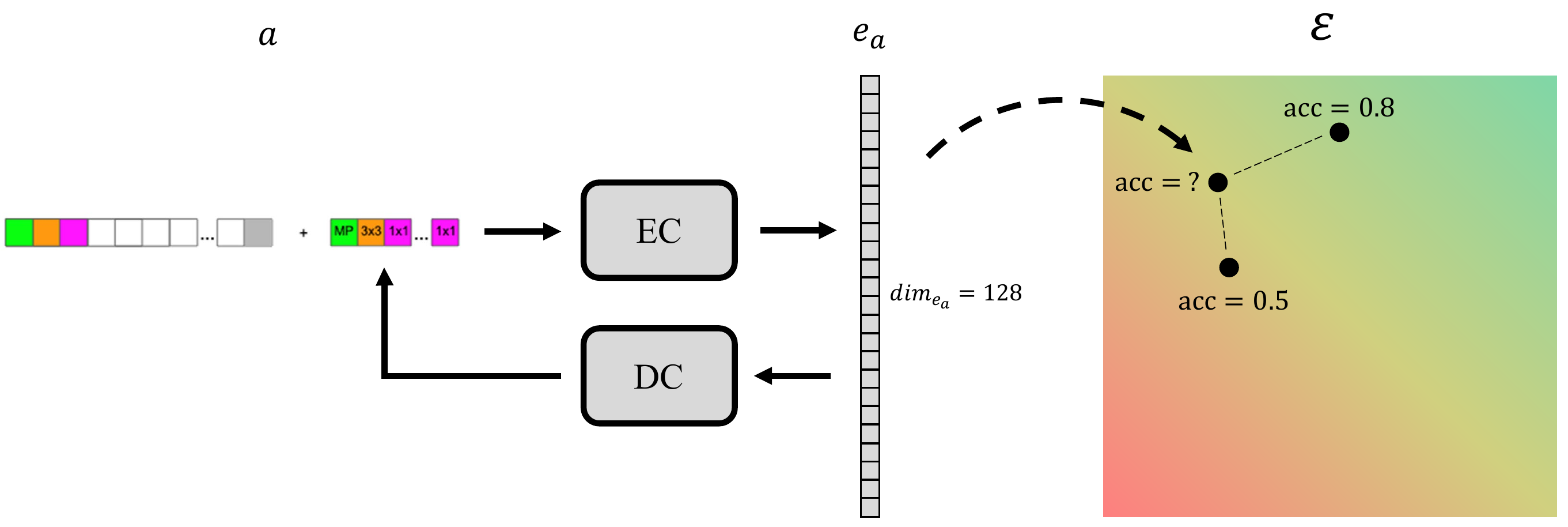}
\end{tabular}
\caption{The EmProx framework. During training, architectures $a$ encoded as a sequence are mapped to embedding space $\epsilon$ using encoder $EC$ to create continuous architecture embeddings $e_a$. Consequently, they are mapped back to the discrete space, their original form $a$, by decoder $DC$ to be able to learn informative embeddings. During inference, the $k$ architectures with known validation accuracy that are the closest to each respective candidate architecture are used to estimate the accuracy of the candidate as weighted average using inverse distance weighting. In the image an example is shown with $k=2$.}
\label{fig:framework}
\end{figure}

EmProx uses an encoder-decoder framework to map candidate architectures to a continuous embedding space. It then estimates the performance of any new architecture based on the distance and accuracy of its nearest neighbors, previously evaluated candidate architectures, in the embedding space. The justification for this method is as follows. NAO creates embeddings of candidate architectures to guide the search and predicts the performance of candidates using an MLP on the embeddings. Training this MLP adds a significant time cost to the search, and would be redundant if we can predict the performance of architectures based one the embeddings in a simpler way. Our proposed method does exactly that: using the embeddings and the accuracy of other architectures the performance of candidates is predicted using kNN, which does not need any training, making the MLP redundant and thus lowering the time cost of the framework. Figure \ref{fig:framework} shows the framework of EmProx and the remainder of this section describes this in more detail.

\subsection{Learning the embedding}
\begin{figure}[htp]
\centering
\begin{tabular}{r}
\includegraphics[height=1.9in]{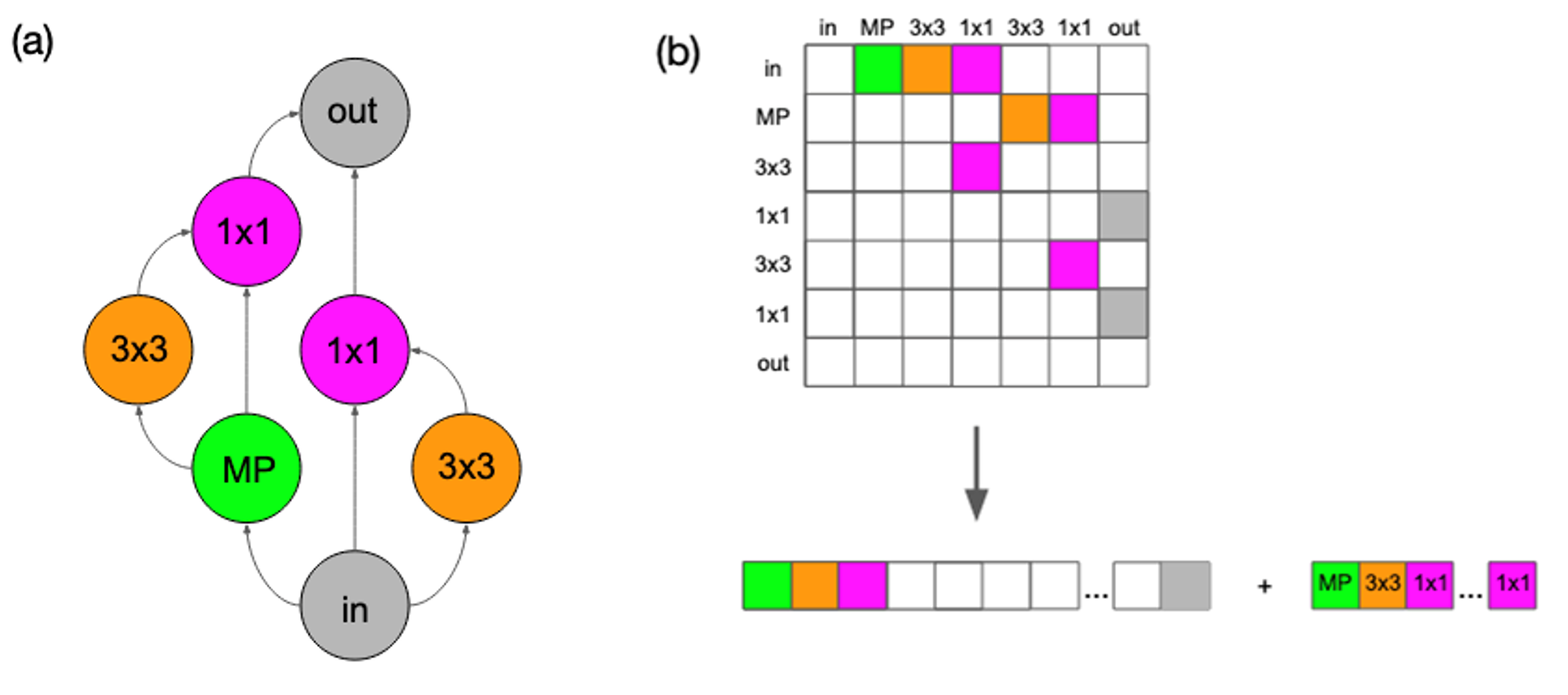}
\end{tabular}
\caption{(a) the way layers and operations in a neural network are connected and (b) the representation as an adjacency matrix and vector.}
\label{fig:encoding}
\end{figure}

Neural network architectures can be represented as a directed acyclic graph. Such graphs can be represented as an adjacency matrix, as shown in Figure \ref{fig:encoding}. Rows and columns represent operations such as convolutions and pooling layers and the values inside the adjacency matrix indicate the existence of a connection between operations in the graph. Consequently, the adjacency matrix can be shown as a sequence, such that a whole architecture can be encoded as a sequence. 

This sequence, describing architecture $a$, serves as input for the framework. At training time, encoder $EC$ maps architecture $a$ to the continuous embedding space $\mathcal{E}$, denoted as $EC: \mathcal{A}\mapsto\mathcal{E}$. Then for all $a \in \mathcal{A}$ we have $e_a = EC(a)$, the embedding of architecture $a$. Given the high dimensionality of the embedding space, the embedding vectors are normalized. Then the decoder $DC$ maps $e_a$ back to its original architecture $a$, denoted as $DC: \mathcal{E}\mapsto\mathcal{A}$. During training, the negative log likelihood loss between the original input and the output of the decoder is minimized. At inference time, to create embedding vectors for the query architectures, the sequences $a$ describing their architectures are fed to encoder $EC$ only.

Both the encoder and decoder consist of a single LSTM layer with a hidden dimension of 128, resulting in an embedding dimension of 128. This dimension was established empirically in the experiments described in section \ref{sec:experiments}. As in the original NAO paper, to better enable the decoder to reconstruct the input, an attention mechanism is used.

\subsection{Performance estimation}
We hypothesize that the encoder-decoder framework maps similar architectures with similar performance closely to each other in the embedding space, such that performance of a new architecture can be inferred from architectures in its vicinity in the embedding space. This is capitalized on as follows. During inference, we have two sets of architectures: one set containing architectures with known accuracy $\mathcal{A}_{known}$ and one set of architectures with unknown accuracy $\mathcal{A}_{unknown}$ for which we want to predict the performance. Both sets are mapped to the embedding space by encoder $EC$ such that we get two new sets. One set containing the embeddings of architectures with known accuracies $E_{\mathcal{A}_{known}} = \{EC(a) \mid a \in \mathcal{A}_{known}\}$ and one set of which we want to predict accuracies $E_{\mathcal{A}_{unknown}} = \{EC(a) \mid a \in \mathcal{A}_{unknown}\}$. \\

We then calculate the distance between all elements in these two sets $\{g(E_{\mathcal{A}_{known}} \times E_{\mathcal{A}_{unknown}})\}$. Here, $g$ denotes the Euclidean distance function, which can be used given that the embedding vectors are normalized. Subsequently, we place the $k$ closest neighboring architectures to each $a \in \mathcal{A}_{unknown}$ in set $N_{a}$. Here, $k$ is a hyperparameter to be chosen later. Then for all $a \in \mathcal{A}_{unknown}$ and neighboring architectures $i = 1 , ..., k$ in $N_{a}$ we calculate the weight $w_i$ based on inverse distance weighting. This is defined as $w_i = \frac{1}{g(e_a, e_i)}$. Lastly, the predicted accuracy $\hat{s}_a$ of architecture $a$ is calculated as the weighted average of the accuracies $s_i$ of neighbors $i = 1 , ..., k$: $\hat{s}_a = \frac{\sum_{i=1}^{k} w_{i}(\mathbf{x}) s_{i}}{\sum_{i=1}^{k} w_{i}(\mathbf{x})}$. \\

\begin{figure}[t]
\centering
\begin{tabular}{r}
\includegraphics[height=1.9in]{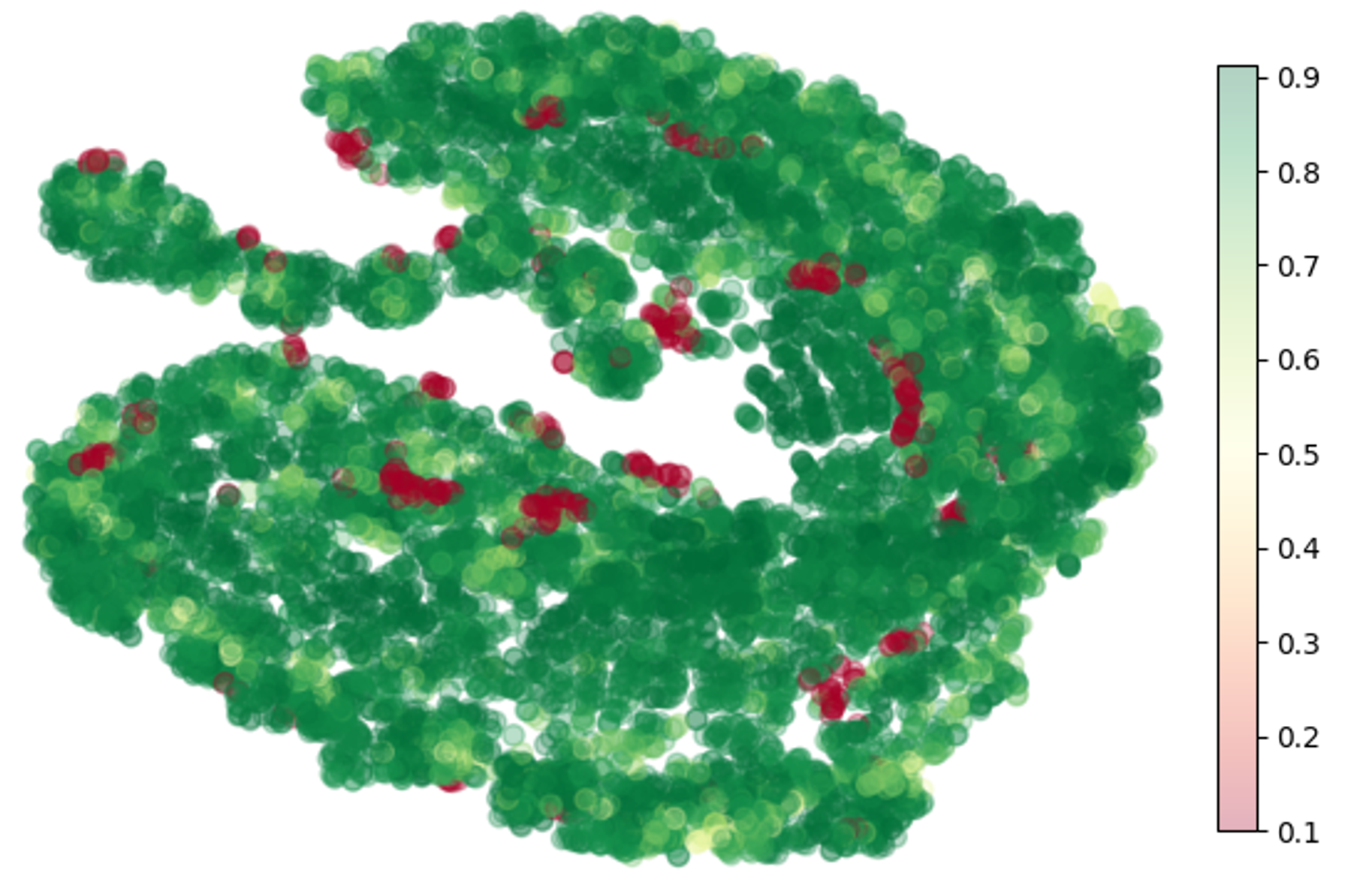}
\end{tabular}
\caption{2D plot of the learned architecture embeddings $e_a$ after reducing the embedding dimesionality to 2 using tSNE. Colors indicate the accuracy of the respective embedded NAS-Bench-201 architectures on the CIFAR10 dataset. The plot shows groups of architectures with similar performance close to each other, which shows that performance of candidate architectures can be estimated from architectures that lie close by in embedding space}
\label{fig:embeddings}
\end{figure}

Figure \ref{fig:embeddings} shows a plot of the learned embeddings, where tSNE was used to reduce the dimensionality to 2. The colors, ranging from red (bad) to green (good), indicate the accuracy of the embedded architectures. As can be seen by the red dotted clusters, several groups of badly performing architectures are grouped together, indicating that the encoder-decoder framework indeed maps similar performing architectures closely to each other in the embedding space.  

\section{Implementation and Evaluation}
\label{sec:design}
To lower the computational barriers for NAS research,  \cite{DBLP:journals/corr/abs-1902-09635} composed NAS-Bench 101, a dataset consisting of 423k unique neural network architectures and their performance evaluated on the CIFAR-10 dataset. This allows for benchmarking NAS algorithms without the necessity to train candidate models from scratch. Likewise, NAS-Bench-201 \citep{DBLP:journals/corr/abs-2001-00326} operates on a cell-based search space, NAS-Bench-301 \citep{DBLP:journals/corr/abs-2008-09777} on DARTS, and NAS-Bench-360 \citep{Tu2021}, on a more diverse set of tasks than CIFAR and Imagenet classification. The strategy proposed in this study is evaluated on NAS-Bench-201 in order to allow the most fair comparison with the strategies evaluated in \cite{White2021} using NASlib.

NASLib \cite{naslib-2020} is a modular framework that implemented many state-of-the-art NAS methods and building blocks like search spaces, search strategies and performance estimation strategies that can be used with only a few lines of code. \cite{White2021} added a total of 31 different neural architecture performance predictor methods to the NASLib framework, and compared them on NAS-Bench-201 with respect to initialization time, query time and performance. 

We evaluate how well our and other performance estimation strategies can predict the actual accuracy of a candidate model in terms of RMSE, MAE and Pearson's correlation coefficient, Spearman's Rho, and Kendall Tau. Based on Pearson and Spearman, \cite{White2021} found that the best methods were Early Stopping (on validation accuracy), GCN \citep{DBLP:journals/corr/abs-1912-00848}, Jacobian Covariance \citep{DBLP:journals/corr/abs-2006-04647}, LcSVR \citep{DBLP:journals/corr/BakerGRN17}, SoTL-E \citep{Ru2020}, SemiNAS \citep{DBLP:journals/corr/abs-2002-10389} and XGBoost \citep{DBLP:journals/corr/ChenG16}. However, not all of these methods could be compared to ours in a fair way. For instance,  Early Stopping is not model-based and could hence not be compared in terms of fit time. Others, such as GCN, far exceeded our resource limitations, which made the comparison infeasible as well as less informative given our focus on efficiency and previously published benchmarks. Hence, to give a fair overview of competitors we chose to benchmark against NAO, SemiNAS, XGBoost and BANANAS \citep{White2019} and an MLP.

To allow a direct comparison, we implemented our method as an addition to NASLib and evaluated it on NAS-Bench-201 architectures in the same manner as in \cite{White2021}, as explained in section \ref{sec:experiments}.

For training and testing, the evaluation framework that is used randomly samples architectures from NAS-Bench-201 until a training set of 121 and a test set of 100 architectures are sampled (random sampling could technically lead to data leakage, though this risk is deemed negligible due to the size of NAS-Bench and the impact of a single architecture being in both the train and test set). This means that 121, a value taken from \cite{White2021}, architectures are used to learn the embeddings.  

\section{Empirical Evaluation}
\label{sec:experiments}
We implemented EmProx in NASLib and experiments were performed using the same framework as \cite{White2021} \footnote{Available at \url{https://github.com/automl/naslib}}. This ensured a fair comparison to other performance estimation strategies implemented in NASLib. In the experiments, the performance predictors of NAO, SemiNAS, XGBoost, BANANAS \citep{White2019} and an MLP were tasked with estimating the performance of NAS-Bench-201 architectures on the CIFAR-10 dataset. In NAS-Bench, the architectures and their validation accuracies are stored. This allowed us to evaluate the performance of the predictors without having to train all architectures on the CIFAR-10 dataset from scratch. As a result, no GPU was required and all experiments have been performed on a single Intel Xeon E5-2698v4 CPU. \\

\begin{table}[htp]
    \begin{adjustwidth}{-1in}{-1in}
        \centering
        \resizebox{1.2\textwidth}{!}{\begin{tabular}{llllllll}
            \toprule
            \multirow{2}{*}{Method} & \multicolumn{2}{c}{Regression metrics} & \multicolumn{3}{c}{Correlation coefficients} & \multicolumn{2}{c}{Time metrics (s)}\\
            \cmidrule(lr){2-3} \cmidrule(lr){4-6} \cmidrule(lr){7-8} \\
            {} & MAE & RMSE & Pearson & Spearman & Kendall & Fit time & Query time \\
            \midrule
            EmProx ($k=10$) & \underline{4.4027 $\pm$ 1.0269} & \underline{10.7163 $\pm$ 2.9690} & 0.4771 $\pm$ 0.1497 & 0.7304 $\pm$ 0.0645 & 0.5453 $\pm$ 0.0537 & \underline{6.4498 $\pm$ 0.9846} & \underline{0.0009 $\pm$ 0.0008} \\
            EmProx ($k=60$) & 4.5264 $\pm$ 0.9625 & 10.7953 $\pm$ 3.1221 & \underline{0.5044 $\pm$ 0.0963} & \underline{0.7332 $\pm$ 0.0865} & \underline{0.5468 $\pm$ 0.0718} & 7.2310 $\pm$ 1.0598 & 0.0032 $\pm$ 0.0004 \\
            NAO & 4.7336 $\pm$ 1.0260 & 10.9394 $\pm$ 2.8279 & 0.4512 $\pm$ 0.1335 & 0.7131 $\pm$ 0.0657 & 0.5279 $\pm$ 0.0608 & 54.1674 $\pm$ 0.3493 & 0.0026 $\pm$ 0.0002 \\
            SemiNAS & \underline{4.0283 $\pm$ 0.8023} & \underline{10.1222 $\pm$ 2.2704} & \underline{0.5307 $\pm$ 0.1604} & \underline{0.7677 $\pm$ 0.0568} & \underline{0.5822 $\pm$ 0.0564} & 152.2268 $\pm$ 42.5809 & 0.0012 $\pm$ 0.0007\\
            XGB & 5.3989 $\pm$ 1.3110 & 12.2955 $\pm$ 3.5906 & 0.4008 $\pm$ 0.2224 & 0.6466 $\pm$ 0.0597 & 0.4719 $\pm$ 0.0476 & \underline{31.6526 $\pm$ 3.9272} & 0.0004 $\pm$ 0.0001 \\
            BANANAS & 7.3007 $\pm$ 1.1988 & 12.2760 $\pm$ 2.0153 & 0.3793 $\pm$ 0.1937 & 0.4169 $\pm$ 0.0959 & 0.2910 $\pm$ 0.0653 & 507.3936 $\pm$ 13.87 & 0.0002 $\pm$ 0.0001\\
            MLP & 6.8584 $\pm$ 1.6152 & 11.3592 $\pm$ 2.2580 & 0.4417 $\pm$ 0.1142 & 0.5298 $\pm$ 0.0996 & 0.3759 $\pm$ 0.0777 & 471.7508 $\pm$ 6.2376 & \underline{$<$0.0001 $\pm$ $<$0.0001}\\
            \bottomrule
        \end{tabular}}

    \end{adjustwidth}
    \caption{Experiment results on the CIFAR10 dataset with NAS-Bench-201 architectures, averaged over 20 trials. Above are the results of the method proposed in this study, below are several methods evaluated in \cite{White2021}. Underlined are the best results of EmProx and the best results among the other predictors for ease of comparison. As can be seen, EmProx scores competitively regarding regression and correlation coefficients and scores much better regarding fit time.}
    \label{tab:results}
\end{table}

As part of the experiments, a grid search was performed to find the dimension of the hidden layer (i.e. embedding dimension) and EmProx hyperparameter $k$. The results of this grid search can be found in the Appendix. Other model and training parameters were left unchanged and taken from \cite{White2021} and the paper of NAO by \cite{Luo2018}. 

The results of the performance predictors were expressed in terms of regression metrics MAE and RMSE between the predicted candidate architecture accuracy and the validation accuracy from NAS-Bench. Additionally, it has to be noted that in many NAS algorithms the exact predicted performance is not of importance, rather than the rank of a certain architecture among other candidates, since only the most promising architectures are trained. This is incorporated in the experiments by including correlation-based performance measures Pearson, Spearman and Kendall's Tau on the predicted and validation accuracies. Furthermore, since ultimately the goal of performance predictors is to speed up the search, fit time and query time are included as measures. For EmProx the fit time includes the learning of the embeddings and the query time the creation of the embedding for a query architecture and the performance estimation. The hyperparameter search was conducted to explore the sensitivity and performance of the method towards its hyperparameters and is hence not included in the fit time.

From the grid search it was found that a hidden layer dimension (i.e. embedding dimension) of 32 yielded the best results. Regarding the number of neighbors $k$, it was found that $k=10$ led to the best results in terms of the MAE, RMSE and fit time. For the rank and correlation coefficients, $k=60$ reached slightly better results. The result on the CIFAR10 dataset using Nas-Bench-201 architectures can be found in Table \ref{tab:results}, alongside with the results of high-performing performance predictors from \cite{White2021} for comparison. All results have been averaged out over 20 trials. 

Compared with the performance predictor of NAO, of which EmProx is a modification, EmProx outperforms NAO on all metrics. This was to be expected for the fit time since EmProx does not need to train the MLP that NAO uses for the performance prediction. The fact that it also outperforms NAO in terms of the regression metrics and correlation coefficients counts as empirical evidence that the method proposed in this study is viable. Regarding the regression and correlation coefficients, SemiNAS yields better results. However, this comes at the cost of the fit time; EmProx is over 20 times faster than SemiNAS. Compared to the second-fastest method, XGBoost, Emprox is still around 5 times faster and performs better on both regression metrics as well as correlation coefficients.

To further validate the performance of EmProx, additional experiments have been conducted. Table \ref{tab:results_cifar100} shows the results of performance prediction of NAS-Bench-201 architectures on the CIFAR100 dataset. This dataset is more complicated than CIFAR10. On the regression metrics and correlation coefficients our method again scores slightly worse than the best predictor from the previous experiments, SemiNAS. However, while on this complicated dataset the fit time of SemiNAS is becoming infeasibly high, EmProx is still outstandingly fast.

Another experiment has been conducted on the DARTS search space. Averaged over twenty trials, EmProx had a RMSE of 0.6652 $\pm$ 0.0694 and a Pearson correlation of 0.4537 $\pm$ 0.0822 with a fit time of 146.1185 $\pm$ 6.9496. These results could not be compared to SemiNAS, since SemiNAS exceeded our resource limitation of 24 hours of CPU time. 

These additional experiments indicate generalization of the findings that our method scores competitively on regression metrics and correlation coefficients with an immensely lower fit time.

\begin{table}[htp]
    \begin{adjustwidth}{-1in}{-1in}
        \centering
        \resizebox{1.2\textwidth}{!}{\begin{tabular}{llllllll}
            \toprule
            \multirow{2}{*}{Method} & \multicolumn{2}{c}{Regression metrics} & \multicolumn{3}{c}{Correlation coefficients} & \multicolumn{2}{c}{Time metrics (s)}\\
            \cmidrule(lr){2-3} \cmidrule(lr){4-6} \cmidrule(lr){7-8} \\
            {} & MAE & RMSE & Pearson & Spearman & Kendall & Fit time & Query time \\
            \midrule
            EmProx ($k=60$) & 5.1859 $\pm$ 0.8155 & 9.914$\pm$ 2.3226 & 0.6212 $\pm$ 0.0792 & 0.7772 $\pm$ 0.0490 & 0.5877 $\pm$ 0.0427 & 16.1493 $\pm$ 0.6764 & 0.0043 $\pm$ 0.0014 \\
            
            SemiNAS & 4.5925 $\pm$ 0.5693 & 9.0315 $\pm$ 1.5319 & 0.6469 $\pm$ 0.1034 & 0.7990 $\pm$ 0.0428 & 0.6124 $\pm$ 0.0418 & 1309.1343 $\pm$ 53.1245 & 0.0108 $\pm$ 0.0049 \\
            \bottomrule
        \end{tabular}}

    \end{adjustwidth}
    \caption{Experiment results on the CIFAR100 dataset with NAS-Bench-201 architectures, averaged over 20 trials. Compared to the best performer in earlier experiments, SemiNAS, our method again scores slightly worse on regression metrics and correlation coefficients, but the fit time is a fraction.}
    \label{tab:results_cifar100}
\end{table}

\section{Conclusion}
\label{sec:conclusions} 
In this study we proposed EmProx, a cheaper and faster method to perform performance estimation during neural architecture search. In the empirical results EmProx was approximately nine times faster than NAO and up to eighty times faster than alternatives, while yielding similarly good predictions according to various regression metrics and correlation coefficients. It also opens up a new area of performance predictors based on proximity in architecture embeddings.

\section{Limitations and Broader Impact Statement}
\label{sec:limitations}
Research on NAS serves the larger societal goal of making machine learning more accessible, and this work specifically speeds up NAS methods, making them more widely applicable. A potential negative impact is found in the consequences of this accessibility; it makes it easier to put machine learning to misuse. 

The results in this study could be improved by better shaping the embedding space, for example by using a variational autoencoder or by adding an extra component in loss the function that promotes architectures with similar performance to be mapped closeby in the embedding space. 

Despite realistic experiments, the proposed method has not been tested as part of a full NAO search. This could further prove the viability of EmProx. Additionally, it has to be noted that EmProx is inherently dependent on the training data, since it bases its predictions directly on the performance of training architectures. Hence, further research should show how sensitive the performance of EmProx is towards the training size. To further show robustness of the method, experiments with resource limits could be conducted, as has been done extensively by \cite{White2021}. Also, currently embeddings are learned from 121 architectures. There is an interesting trade-off between embedding quality and fit time as a result of this choice of number of architectures that could further be explored. Lastly, it would be interesting to see how well the performance of this method generalizes to more complex datasets such as CIFAR-100 or ImageNet and more complex search spaces like NAS-Bench301. 

%

%

\newpage
\bibliography{main}



\newpage
\section*{Appendix}
\subsection{Hyperparameter search}
For hyperparameter tuning, a gridsearch was performed on the number of neighbors $k=[3, 10, 60]$ and the dimension of the embedding $d=[16, 32, 64, 128]$. These specific values for $k$ are the result of earlier exploratory analysis. The results can be found in table \ref{tab:gridsearch}. By no means this grid search aimed to find the most optimal configuration of EmProx. It merely served as an exploration of the performance of the method and its sensitivity towards its hyperparameters.

\begin{table}[htp]
    \centering
    \begin{tabular}{lllllllll}
        \toprule
        \multicolumn{2}{l}{ } & \multicolumn{2}{c}{Regression metrics} & \multicolumn{3}{c}{Correlation coefficients} & \multicolumn{2}{c}{Time metrics (s)}\\
        \cmidrule(lr){3-4} \cmidrule(lr){5-7} \cmidrule(l){8-9} \\
        k & d & MAE & RMSE & Pearson & Spearman & Kendall & Fit time & Query time \\
        \midrule
        3 & 16 & 5.0856 & 11.7021 & 0.3918 & 0.6134 & 0.4416 & 8.6227 & 0.0006 \\
        3 & 32 & 4.7855 & 11.4074 & 0.4003 & 0.6834 & 0.4968 & 6.4858 & 0.0007 \\
        3 & 64 & 4.9680 & 11.5436 & 0.3924 & 0.6656 & 0.4863 & 7.4364 & 0.0008 \\
        3 & 128 & 4.9231 & 11.5629 & 0.4136 & 0.6805 & 0.5011 & 26.0462 & 0.0034 \\
        10 & 16 & 4.7314 & 11.0031 & 0.4614 & 0.6744 & 0.4966 & 6.1812 & 0.0008 \\
        10 & 32 & 4.4027 & 10.7163 & 0.4771 & 0.7304 & 0.5453 & 6.4498 & 0.0009 \\
        10 & 64 & 4.6388 & 10.7278 & 0.4667 & 0.6991 & 0.5178 & 7.3694 & 0.0010 \\
        10 & 128 & 4.6323 & 10.8578 & 0.4551 & 0.7200 & 0.5331 & 26.0007 & 0.0044 \\
        60 & 16 & 4.7076 & 11.0466 & 0.4808 & 0.7095 & 0.5246 & 6.5411 & 0.0032 \\
        60 & 32 & 4.5265 & 10.7953 & 0.5044 & 0.7332 & 0.5468 & 7.2310 & 0.0032 \\
        60 & 64 & 4.7681 & 10.9440 & 0.4809 & 0.7083 & 0.5247 & 8.4264 & 0.0035 \\
        60 & 128 & 4.6267 & 10.8773 & 0.4779 & 0.7209 & 0.5369 & 17.2969 & 0.0053 \\
        \bottomrule
    \end{tabular}
    \vspace*{5mm}
    \caption{Gridsearch results over parameters $k$, the number of neighbors to base the performance estimation on, and $d$, the dimension of the hidden layer/embedding. Results are averaged over 20 trials.}
    \label{tab:gridsearch}
\end{table}

\end{document}